\documentclass[runningheads]{llncs}
\usepackage[pdftex]{graphicx}
\usepackage{xcolor,colortbl,times,soul,url,booktabs,todonotes,tabularx,ifthen,booktabs,amsmath,hyperref,cleveref, url, floatrow, blindtext}

\usepackage{tikz}
\usetikzlibrary{arrows,shapes,positioning,shadows,trees}
\tikzset{
  basic/.style = {draw, text width=3cm, font=\sffamily, rectangle, rounded corners=6pt, thin,align=center},
  root/.style   = {basic,fill=white!30},
  level 3/.style = {basic, fill=yellow!30},
  level 4/.style = {basic, fill=pink!60},
  level 5/.style = {basic, fill=orange!30},
}
\usepackage[utf8]{inputenc}
\usepackage[graphicx]{realboxes}
\newcommand{\etal}{et. al.,}
\newcommand{\scs}{SCS}

\titlerunning{Survey of Machine Learning and Safety}

\author{Mehrnoosh Askarpour
\orcidID{0000-0001-6526-2544} 
\and
Alan Wassyng
\orcidID{0000-0003-4614-3421} 
\and 
Mark Lawford
\orcidID{0000-0003-3161-2176} 
\and
Richard Paige
\orcidID{0000-0002-1978-9852}
\and
Zinovy Diskin
\orcidID{0000-0001-8025-4630} 
}
\authorrunning{Askarpour et al.}
\institute{McMaster University, Hamilton, Canada\\
\email{\{askarpom,diskinz,lawford,paigeri,wassyng\}@mcmaster.ca
}\\
}
\begin{document}
\title{Is the Rush to Machine Learning Jeopardizing Safety? Results of a Survey}
\maketitle
\begin{abstract}
Machine learning (ML) is finding its way into safety-critical systems (\scs). Current safety standards and practice were not designed to cope with ML techniques, and it is difficult to be confident that \scs s that contain ML components are safe.
Our hypothesis was that there has been a rush to deploy ML techniques at the expense of a thorough examination as to whether the use of ML techniques introduces safety problems that we are not yet adequately able to detect and mitigate against. We thus conducted a targeted literature survey to determine the research effort that has been expended in applying ML to \scs\ compared with that spent on evaluating the safety of \scs s that deploy ML components. This paper presents the (surprising) results of the survey.
\end{abstract}
\section{Introduction}
\label{sec:int}
Machine learning (ML) is rapidly being incorporated in the design of safety-critical systems (\scs)~\cite{8893310} for many different purposes such as object recognition, computer vision, and navigation. The ability to solve complex problems while improving performance is a primary reason for the prevalent and ever-growing use of ML.

Working in the safety domain, we have an uneasy feeling about this rapid shift, since the current practices and regulations in various domains (such as ISO~26262 for road vehicles and ISO~10218 for robotic systems) are not adequate to tackle the complexities and data-driven nature of ML based systems~\cite{8987559}, as they are not developed based on requirements, design trace-ability or functional needs~\cite{jenn2020identifying}.
We started by exploring the existing literature on safety and ML. Our initial findings escalated our uneasiness, because we inferred that more research was done on integrating ML in safety-critical systems (\scs) compared with assessing the safety of those integrated components. Thus, we decided to perform a targeted literature review to determine whether or not this is true. This paper reports the results of our literature review.

Our evaluation of the current state of the art suggests three main research directions in the domain of ML and \scs s: (C1) ML for the design of \scs s, (C2) the safety of systems that contain ML components, and (C3) using ML techniques to analyse the safety of \scs s that may or may not embed ML components. 
The three classes are not strictly separable and overlap in different ways.
In particular, C1 and C3 are similar in that they both discuss the application of ML in \scs s, while C2 is about coping with ML components that are already embedded in \scs s. In our analysis we found papers that, regardless of the super/sub class relationships of the three classes, tackle the issues that are covered in more than one of them (see Table 3.)

We provide statistical analysis on the current research directions and compare the dedicated attention to C1, C2, and C3. In particular -- and this may be seen as a limitation of this research -- we chose a set of relevant venues as a statistical population and investigated the number of publications in each of them (see \Cref{tab:venues}). We hoped to see, despite our intuition, that the same amount of effort applied to all of them.
%
%
\begin{table}[!htb]
\caption{Targeting publication venues. {We evaluated \href{https://dblp.org/db/conf/ecmdafa/index.html}{ECMFA}, other  \href{https://dblp.org/db/conf/icse/index.html}{ICSE} tracks, and \href{https://dblp.org/db/conf/ease/index.html}{EASE} but did not find relevant papers.} {ICMLA 2020 proceedings are not available as up to Feb. 2020.}} 
{
\begin{tabular}{l|l}
\toprule
	Acronym & {Venue Name}\\
	\midrule
	\multicolumn{2}{c}{\textbf{Safety}}\\
	\midrule
	SAFECOMP & {\centering Int. Conference on Computer Safety, Reliability, and Security}\\
	{ISSRE} & {IEEE Int. Symposium on Software Reliability Engineering}\\
	RESS  & Reliability Engineering \& System Safety Journal\\
	ICVES & Int. Conference on Vehicular Electronics and Safety \\
	SS & Journal of Safety Science \\
	\midrule
	\multicolumn{2}{c}{\textbf{Software Engineering and Model-Driven Engineering}}\\
	\midrule
	MODELS & Int. Conference on Model Driven Engineering Languages and Systems\\
	JSS & Journal of Systems and Software \\
	ASE & Automated Software Engineering\\
	TSE & IEEE Transactions on Software Engineering\\
	ICSE - SEAMS & Int. Symp. on Software Engineering for Adaptive and Self-Managing Systems\\
	\midrule
	\multicolumn{2}{c}{\textbf{Machine Learning}}\\
	\midrule
	ICML & Int. Conference on Machine Learning \\
	ICMLA & Int. Conference on Machine Learning and Applications  \\
	IJCAI & Int. Joint Conference on Artificial Intelligence \\
	NeurIPS & Neural Information Processing Systems\\
	ICLR & Int. Conference on Learning Representations\\
	LOD  & Int. Conference on Machine Learning, Optimization, and Data Science\\
	\midrule
	\multicolumn{2}{c}{\textbf{Transportation}}\\
	\midrule
	IV & IEEE Intelligent Vehicles Symposium \\
	\midrule
	ITSC & Int. Conference on Intelligent Transportation Systems\\
	\bottomrule 
\end{tabular}	
}
\label{tab:venues}
\end{table}
%
\section{Research Method}
\label{sec:meth}
\begin{figure}[!htb]
\centering
\scriptsize{
\begin{tikzpicture}[
  level 1/.style={sibling distance=40mm},
  edge from parent/.style={->,draw},
  >=latex,level distance=22pt]
\node[root] {ML, Safety and Reliability}
  child {node[root] (c0) {(C1) ML for \scs ~design}}
  child {node[root] (c2) {(C2) Reliability and safety of ML systems}}
  child {node[root] (c1) {(C3) ML for reliability and safety applications}}
;
\begin{scope}[every node/.style={level 3}]
\node [below = .05 cm of c0, xshift=15pt] (c01) {Creating system models};
\node [below = .05 cm of c01] (c02) {Creating component models};
\node [below = .05 cm of c02] (c03) {Extracting uncertainty parameters };
\node [below =  .05 cm of c03] (c04) {Draw out safety constraints };
\end{scope}
\begin{scope}[every node/.style={level 5}]
\node [below = .05 cm of c1, xshift=15pt] (c11) {Regression};
\node [below = .05 cm of c11] (c12) {Classification};
\node [below =  .05 cm of c12] (c13) {Clustering};
\node [below =  .05 cm of c13] (c14) {Anomaly detection};
\node [below = .05 cm of c14] (c15) {Semi-supervised learning};
\end{scope}
\begin{scope}[every node/.style={level 4}]
\node [below = .05 cm of c2,xshift=15pt] (c21) {Size and representativeness of the training dataset};
\node [below = .05 cm of c21] (c22) {Feature selection};
\node [below = .05 cm of c22] (c23) {Over-fitting problem};
\node [below = .05 cm of c23] (c24) {Mismatches between the model and the environment behaviour};
\end{scope}
\foreach \value in {1,2,3,4} \draw[->] (c0.180) |- (c0\value.west);
\foreach \value in {1,2,3,4,5} \draw[->] (c1.180) |- (c1\value.west);
\foreach \value in {1,2,3,4} \draw[->] (c2.180) |- (c2\value.west);
\end{tikzpicture}
}
\caption{The ontology of the interdisciplinary area of ML and Safety.}
\label{fig:ontology}
\end{figure}
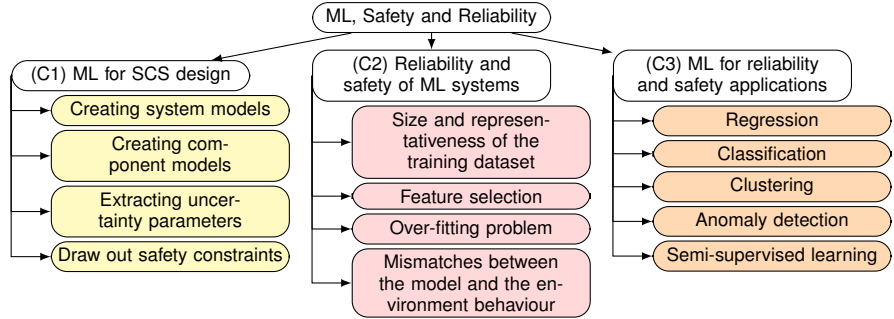

We first precisely defined our research questions, then we defined a methodology to explore the state of the art with a predefined paper inclusion criteria. 
\\\textbf{\textit{Research Questions:}} 
Our main research questions are: (\textbf{RQ1}) How many papers discuss C1, C2, and C3 respectively? (\textbf{RQ2}) Is there a large gap between the answers to RQ1? and (\textbf{RQ3}) How have each of the three directions grown or shrunk over the years, and is a larger gap between them plausible in the near future?
\\\textbf{\textit{Paper Selection and Criteria:}} 
Typically, systemic literature reviews conduct a query with certain keywords over one of the reliable academic research databases such as Google Scholar or Scopus. However, we saw one big issue in proceeding as such, and that was choosing a comprehensive enough set of keywords; We could have searched for combinations of ``safety" and ``machine learning" in titles and abstracts of papers, however the ML domain has several sub-domains (e.g., NN, DL,...) and simply searching for ML could lead to missing out on papers that directly discuss those sub-domains. Besides, listing all the keywords that reflect the sub-domains of ML might be a challenge. So, we decided to select a collection of high quality venues in ML, safety and software engineering domains and to inspect their published papers in the last six years (2015 - 2020).
We manually examined the proceedings and volumes of the selected venues listed in \Cref{tab:venues}, and analyzed the abstract and introduction section of any paper with a title that contained one of the following keywords: 
{\footnotesize
\begin{center}
\textit{(``Safety" OR ``Certification" OR ``Assurance" OR ``Risk")
 AND 
(Any ML Keyword including ``Bayesian Networks", ``Deep learning", ``Supervised learning", ``Markov Process")}
\end{center}
}
We suspected that there might be relevant work that does not contain ML keywords in their title, abstract or keywords, so we added those we knew of to the set of our search results, even if they were not a result of our query. Note that we chose venues of \Cref{tab:venues} such that our results would not reflect a single domain. The last two venues were not included in our first round of exploration. Later, in \Cref{sec:res}, we explain why we added two domain specific venues and what we obtained by doing that.
\\\textbf{\textit{Data Extraction and Analysis:}} 
We manually analyzed the proceedings and volumes of the venues listed in \Cref{tab:venues} by searching our query in DBLP and journal homepages. After pruning them based on title and abstraction, we further pruned them by reading their introduction sections. We then fully scanned the remaining papers to confirm their relevance.
\Cref{fig:method} shows our process steps, and the number of papers analysed in each step, excluding key-notes, posters, invited talks, oral presentations, and reviews.
\\\textbf{\textit{Ontology:}} 
\label{sec:ont}
We visualize the ontology of the domain in \Cref{fig:ontology}; we envision using this as part of a road-map for possible future research directions.
The yellow boxes depict ML used in \scs~ design. ML algorithms use the execution log histories, accident documentation, performance reports, etc.,  as the training dataset to create models of the system or components within the system, and to extract uncertainty parameters or draw safety constraints.
The red boxes, inspired by Faria \etal~\cite{faria2018machine}, reflect the challenges of ML that potentially affect the safety and reliability of ML systems. Qualifications of the training dataset (i.e., its size, validity, distribution and distributional shift) and how feature selection is performed on it, notoriously affect the driven model and constraints, and might cause over-fitting.
Moreover, any mismatch between the generalization and the reality could lead to safety cracks directly or indirectly (i.e., generalizing inadequate policies or inefficient safety values for safe exploration).
The orange boxes show which safety analysis processes could be implemented via ML techniques~\cite{DBLP:journals/corr/abs-2008-08221}. 
%
\section{Search Results}
\label{sec:res}
\begin{figure}[!htb]
\begin{floatrow}
\ffigbox{%
\includegraphics[width=.45\textwidth]{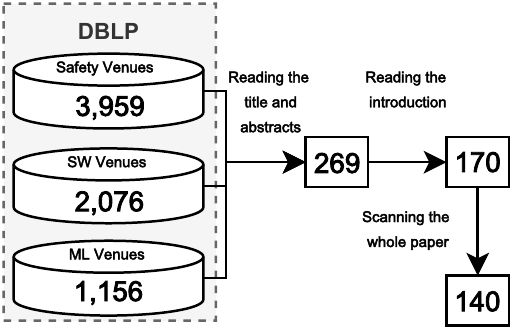}}
   { \caption{The number of papers filtered out during different phases of the adopted research method. The numbers exclude venues IV and ITCS.}
    \label{fig:method}
}
\capbtabbox{%
 {
 \begin{tabularx}{.5\textwidth}{l | l}
\toprule
	Community & Order of focus \\
	\midrule
	{ Safety} & { C3(51) $>$ C2(15) $>$ C1(14)}\\
	{ Software Eng.} & { C3(14)  $>$ C2(7)  $>$ C1(4)}\\
	{ ML} & { C1(23) $>$ C2(19) $>$ C3(11)}\\
	\bottomrule 
\end{tabularx}	
}
\label{Tab:focus}
}{%
  \caption{The order of focus of each community.}
}
\end{floatrow}
\end{figure}

\begin{figure}[!htb]
\begin{floatrow}
\ffigbox{%
  \includegraphics[width=.45\textwidth]{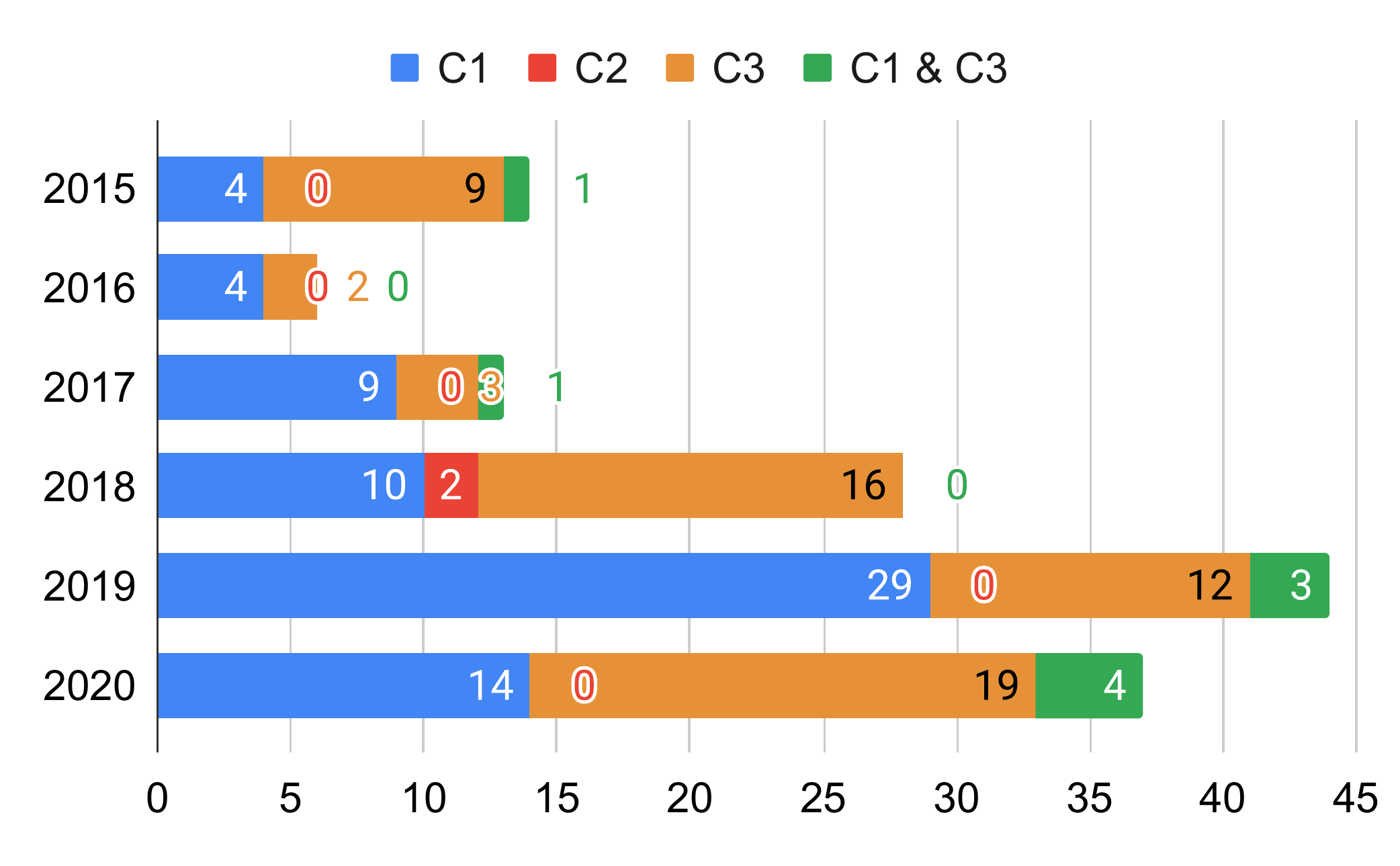}
}{%
 \caption{Results of our search in ITSC.}
\label{fig:itsc}
}
\ffigbox{%
\includegraphics[width=.45\textwidth]{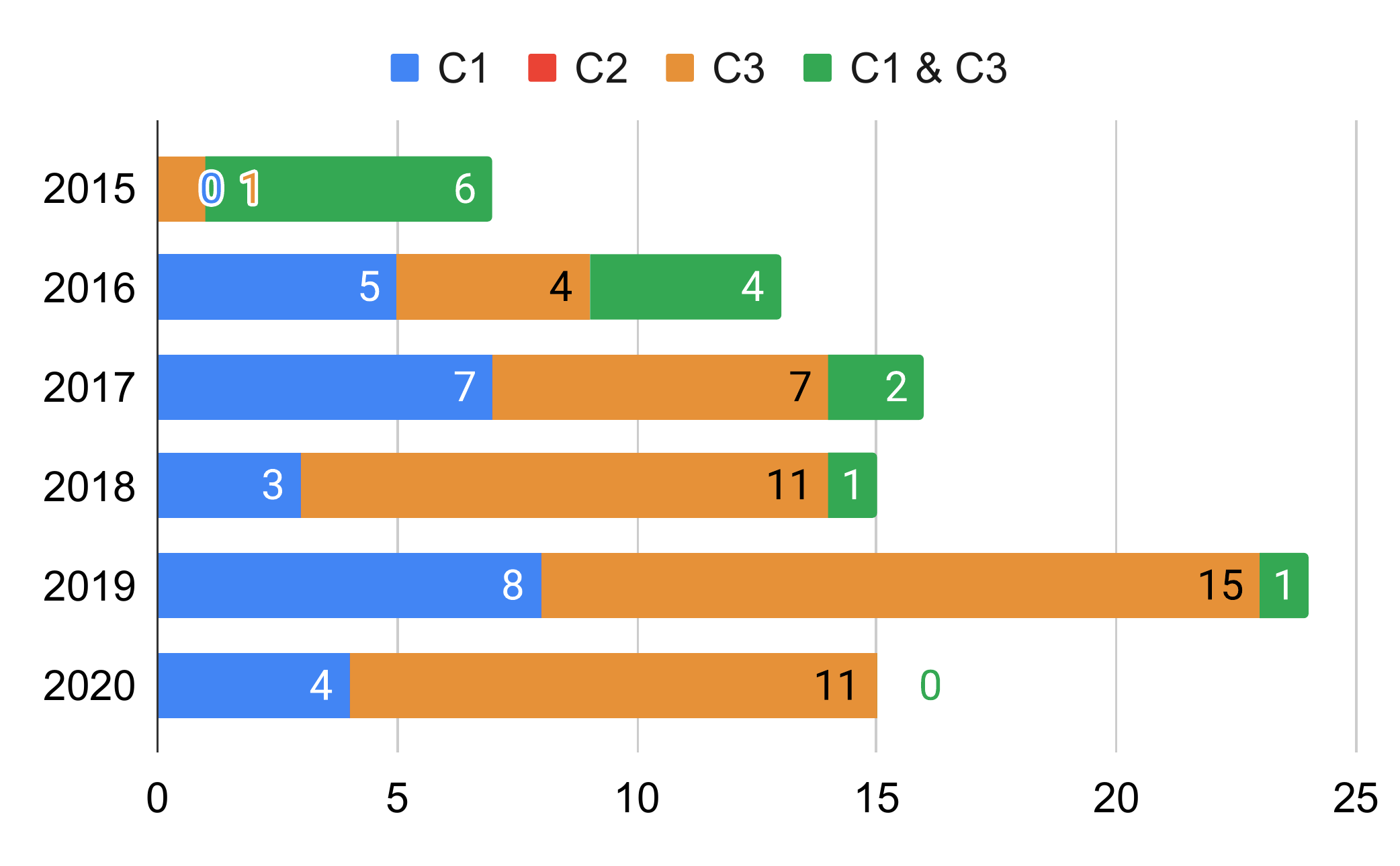}}
{\caption{Results of our search in IV.}
\label{fig:iv}
}
\end{floatrow}
\end{figure}

After applying the methodology explained in \Cref{sec:meth}, we found a total of 140 relevant papers (\Cref{fig:method}), created Table 3 and found the following answers to our questions:
\newfloatcommand{capbtabbox}{table}[][\FBwidth]
\\\textbf{\textit{RQ1:}} There were 38, 36, and 74 papers for C1, C2, and C3 respectively.
\\\textbf{\textit{RQ2:}}
The results show a large focus on C3. Note that the number of papers for C3 is almost twice that of C1 or C2 (see Table 3), which means that ML is more often used as an external tool to analyze safety rather than as a safety-critical component. 
Despite the possible relationship between C2 and C3, there is not a single work that treats them both.
Surprisingly, we do not see a large gap between C1 and C2. This is good news -- however too good to be true. We suspected that the reason that we did not find more C1 papers with our query is that many papers discuss the use of ML in a system that is potentially safety critical, or part of a safety critical system, but the main focus of the paper is not safety. Hence, they do not even use ``safety-critical" to describe their system of interest.  To verify this theory, we decided to pick a couple of venues that are particular to safety critical systems and search only for ML keywords in them with the justification that the importance of ``safety" is inherent in those works. We chose IEEE Intelligent Vehicles Symposium (IV) and International Conference on Intelligent Transportation Systems (ITSC) for this purpose. This time, we were not surprised to see that there is a large gap between C1 and C2. In fact, we did not even find one instance of C2 in the last six years of IV, and detected only a couple in proceedings of ITSC. We provided the numbers of papers found in IV and ITSC in \Cref{fig:iv} and \Cref{fig:itsc} respectively but could not include all the citations due to the space limit.

In the light of this observation and also the results shown in Table 2, we can conclude that: i) there is a large gap between C1 and C2 at least in transportation and autonomous vehicles domains; ii) the same  gap might exist in several other safety critical domains which requires a different research scope from the one we chose (i.e., more concentrated on one single domain); iii) ML and software engineering research communities tend to ignore the impact of ML on safety; and iv) the safety community tends to be more cautious about the safety of ML systems, yet focuses more on C3 rather than C2.

\begin{figure}[!htb]
\begin{floatrow}
\ffigbox{%
  \includegraphics[width=.45\textwidth]{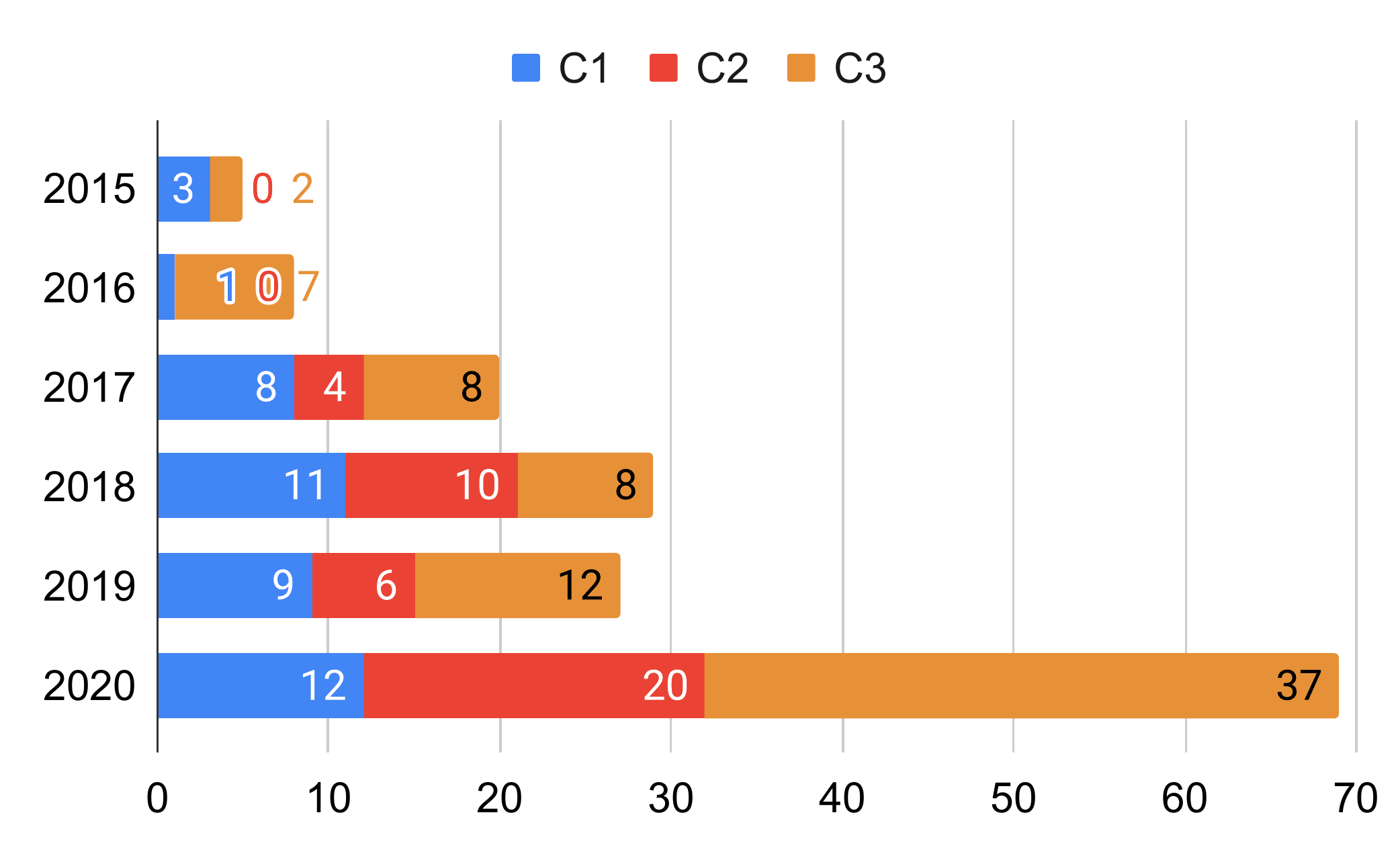}
}{%
  \caption{Numbers of papers regarding C1 (blue), C2 (red), and C3 (yellow) per year.}
         \label{fig:year-pie}
}
\ffigbox{%
\includegraphics[width=.45\textwidth,trim={3cm .5cm 3cm 0.1cm },clip]{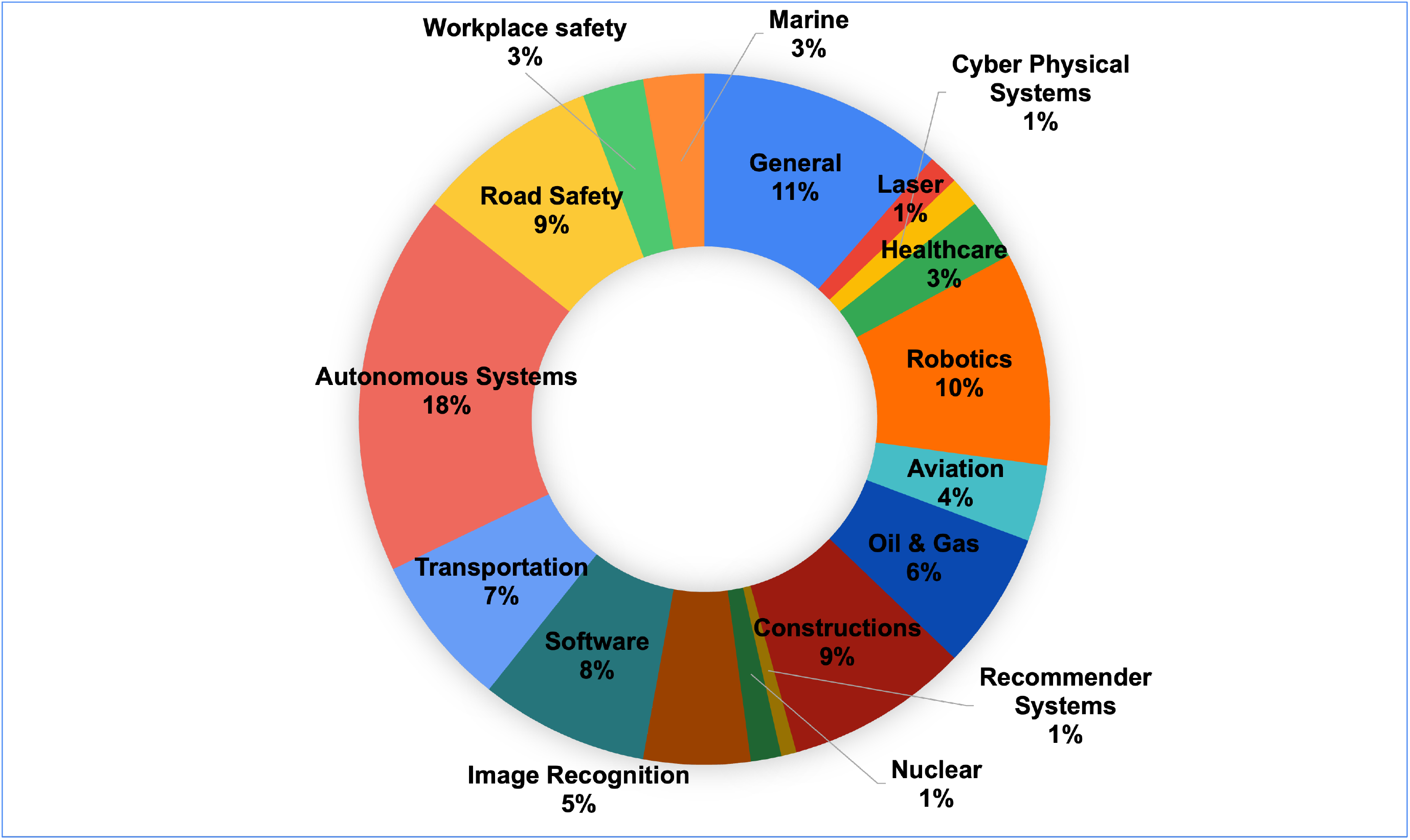}}
{\caption{Distribution of results of our search over various application domains.}
\label{fig:app-pie}
}
\end{floatrow}
\end{figure}
\textbf{\textit{RQ3:}} \Cref{fig:year-pie} shows the number of papers about each research direction per year. It is evident that the focus on C3 has been increasing, while the number of C1 and C2 papers has fluctuated over the last six years. \Cref{fig:year-pie} shows that there is still a gap between C1 and C2, but it is decreasing over time, indicating that the importance of C2 is being recognized by the community. In particular, in 2020 there was a rise in publications discussing C2.
\Cref{Tab:focus} also shows that C2 is the second focus of the three communities and has been the focus of more papers than C1.
However, as discussed for RQ2, \Cref{fig:iv} and \Cref{fig:itsc}, which are domain specific venues, draw a different picture of the situation where C2 hardly appears. 
As a further observation, the prevalence of the targeted application domains impacted the most by ML are shown in \Cref{fig:app-pie}.

%
\begin{table}[!htb]
\begin{minipage}{\linewidth}
{\scriptsize
\begin{tabularx}{\textwidth}{l|l|X}
\toprule
\multicolumn{3}{c}{\textbf{Safety}}\\\midrule
SAFECOMP & C3 & \cite{DBLP:conf/safecomp/MartinelliMNSV18,DBLP:conf/safecomp/CzarneckiS18},\cite{DBLP:conf/safecomp/ZhaoBSRF0020,10.1007/978-3-030-54549-9_13} \\
 & C2 & \cite{10.1007/978-3-319-66284-8_1},\cite{10.1007/978-3-319-99130-6_4,10.1007/978-3-319-99229-7_2,10.1007/978-3-319-99229-7_35},\cite{10.1007/978-3-030-26601-1_12,10.1007/978-3-030-26250-1_30,10.1007/978-3-030-26250-1_33}, \cite{10.1007/978-3-030-55583-2_30,10.1007/978-3-030-55583-2_29,10.1007/978-3-030-55583-2_28,10.1007/978-3-030-55583-2_25,10.1007/978-3-030-55583-2_24}\\
 & C1 & \cite{10.1007/978-3-319-99229-7_39,10.1007/978-3-319-99229-7_36,10.1007/978-3-319-99229-7_48}\\\midrule
 ISSRE & C2 & \cite{9307660}\\
       & C3 & \cite{ceabc3b89afc458ab6b979573771641c}\\\midrule
 RESS   & C3 & \cite{WU2015157},\cite{ZHAO201774},\cite{ZHOU2018152,IAMSUMANG2018118},\cite{STEIJN2020104514,YANG2020104437,FANG2020104604,LI2020104655,SARKAR2020104616,ZHAO2020104888,KURIAN2020104873,SARKAR2020104900,ARTEAGA2020104988,HUANG2020107220,XU2020107210,FAN2020107070,ZHANG2020107089,ZHAO2020106878,KIM2020106959,KIM2020106959,ZHOU2020106902,MEANGO2020106827,RUIZCASTRO2020106797,UTNE2020106757,WANG2020106781,WANG2020106705,RAHMAN2020106623,DUAN2020106676,DHULIPALA2020106659,TALEBBERROUANE2020106587}\\
        & C2 & \cite{WU2020107051}\\
        & C1 & \cite{DBLP:journals/ress/PiriouFL17},\cite{ZHANG2020107094}\\
        & C1 \& C3 & \cite{KHAN2020104858}\\
        \midrule
ICVES & C3 & \cite{8906396}\\
      & C1 & \cite{7991914},\cite{8519504}\\
      \midrule
SS  & C3 & \cite{DAI201556},\cite{JITWASINKUL2016264},\cite{PATRIARCA201749},\cite{WANG2019833,LIU2019764,CHEN2019268,PENG2019263,YU2019248,WASHINGTON2019654,LIANG2019861,YANG2019205,PALTRINIERI2019475}\\
    & C1 & \cite{KALININA2018164,DINDAR201820,KONDO2018225}, \cite{FAN2019607,BI2019435,OEHLING201989}\\
	\bottomrule 
\end{tabularx}	}
	\label{tab:sur}
 \end{minipage}%
\hspace{1cm}

\begin{minipage}{0.3\linewidth}
{\scriptsize\begin{tabularx}{\textwidth}{l|l|X}
\toprule
	\multicolumn{3}{c}{\textbf{Software Engineering}}\\\midrule
	MODELS & C1 & \cite{DBLP:conf/models/SchoneMRA19},\cite{DBLP:journals/sosym/0001MFT19}\\\midrule
    JSS &  C2 & \cite{DBLP:journals/jss/MostaeenRRSS20},\cite{DBLP:journals/jss/BraiekK20}\\
        & C3 & \cite{YU201744}, \cite{KUMAR2018686}, \cite{DBLP:journals/jss/MartinVANB20}\\
        & C3 \& C1 & \cite{DBLP:journals/jss/AnLZCCS20,DBLP:journals/jss/BarbezKG20}\\
        \midrule
    ASE & C2 & \cite{DBLP:conf/kbse/Abdelkader20,DBLP:conf/kbse/Kwiatkowska20}\\
        & C3 & \cite{DBLP:journals/ase/WangZJZ16},\cite{DBLP:journals/ase/ZhangJW17}, \cite{DBLP:conf/kbse/Masalimov20}\\
        & C3 \& C1 & \cite{10.1145/3243127.3243129} \\\midrule
    TSE & C3 & \cite{7450669,DBLP:journals/tse/Tantithamthavorn16},\cite{7990255}, \cite{8359087}\\\midrule
    SEAMS & C2 & \cite{DBLP:conf/icse/ScheererKRB20,DBLP:conf/icse/PasareanuCFG20}\\
          & C3 & \cite{10.1145/3194133.3194147}\\
	\bottomrule 
\end{tabularx}	}
 \end{minipage}%
\hspace{1.5cm}
 \begin{minipage}{0.5\linewidth}
  \label{tab:prices2012}
  \centering
{\scriptsize\begin{tabularx}{\textwidth}{l|l|X}
\toprule
	\multicolumn{3}{c}{\textbf{Machine Learning}}\\\midrule  
	ICML & C1 & \cite{DBLP:conf/icml/SuiGBK15,DBLP:conf/icml/Bou-AmmarTE15},\cite{DBLP:conf/icml/SunDK17},\cite{DBLP:conf/icml/KirschnerMHI019},
	\cite{DBLP:conf/icml/GuoZJLZ20,DBLP:conf/icml/BrownCSN20,DBLP:conf/icml/BaoGH20,DBLP:conf/icml/AtamturkG20} \\
	\midrule
	ICMLA  & C3 \& C1 & \cite{8260754}\\
	       & C1 & \cite{7838199}\\
	\midrule
	IJCAI & C3 & \cite{DBLP:conf/ijcai/ZhangYCAB16},\cite{DBLP:conf/ijcai/McAllisterGKWSC17,DBLP:conf/ijcai/FreedBHSB16},\cite{DBLP:conf/ijcai/McDermidJH19,DBLP:conf/ijcai/HeinzmannSOSK19} \\
	      & C2 & \cite{DBLP:conf/ijcai/ZhangDS18},\cite{DBLP:conf/ijcai/KrakovnaOML19,DBLP:conf/ijcai/Wotawa19}\\
	      & C1 & \cite{DBLP:conf/ijcai/MansouriLHP19,DBLP:conf/ijcai/TurnerHT19,DBLP:conf/ijcai/MancusoKLS19}\\      
	      & C3 \& C1 & \cite{DBLP:conf/ijcai/GodoyKGG16},\cite{DBLP:conf/ijcai/SternJ17}\\
	\midrule
	NeurIPS & C2 & \cite{DBLP:conf/nips/MhamdiGHM17},\cite{DBLP:conf/nips/WangPWYJ18,DBLP:conf/nips/WenT18,DBLP:conf/nips/ChowNDG18,DBLP:conf/nips/HuangWP018,DBLP:conf/nips/ZimmerMN18,DBLP:conf/nips/TsuzukuSS18} \\
	        & C2 \& C1 &  \cite{DBLP:conf/nips/BerkenkampTS017,DBLP:conf/nips/TurchettaB016}\\
	\midrule
	ICLR & C2 & \cite{DBLP:conf/iclr/SinghGPV19}\\
	     & C1 & \cite{DBLP:conf/iclr/SinhaND18,DBLP:conf/iclr/RaghunathanSL18}\\
	     & C3 \& C1 & \cite{DBLP:conf/iclr/EysenbachGIL18}\\
	\midrule
	LOD & C1 & \cite{DBLP:conf/nips/ChandakJTWT20} \\
    	& C3 & \cite{DBLP:conf/mod/LazzeriniP17}\\
	    & C2 & \cite{DBLP:conf/mod/AndersenGG20}, \cite{DBLP:conf/nips/TurchettaKS0A20,DBLP:conf/nips/SinhaOTD20} \\
	    & C3 \& C1 & \cite{DBLP:conf/nips/LuoSK20}\\
	\bottomrule 
\end{tabularx}	}
 \end{minipage}%
 \caption{The list of venues and 140 papers found over 2015-20, excluding IV and ITSC.}
\end{table}
%
\section{Conclusions}
\label{sec:conc}
This paper provides a targeted literature survey and an analysis of the past six years of published research related to ML and \scs . We targeted highly regarded and relevant conferences and journals to reduce the scope of the survey, but also to be able to make statements about how various research communities have fared regarding ML and \scs . Our original idea was to draw attention to the lack of research on the safety of \scs s that rely on ML components. An interesting outcome of the survey is that our original hypothesis would have been true prior to 2020. It is encouraging that in many application domains, research related to safety of \scs s with embedded ML components is now getting the attention it should. A disconcerting result is that in the automotive domain, if our selected publication venues are representative, almost all the effort is going into ML usage in design of \scs s or in evaluation of safety using ML technology.
\bibliographystyle{splncs04}
\bibliography{bib.bib}

\begin{thebibliography}{100}
\providecommand{\url}[1]{\texttt{#1}}
\providecommand{\urlprefix}{URL }
\providecommand{\doi}[1]{https://doi.org/#1}

\bibitem{DBLP:conf/kbse/2020}
35th {IEEE/ACM} Int. Conf. on Automated Software Eng., {ASE}. {IEEE} (2020)

\bibitem{DBLP:conf/icml/2020}
Proc. of the 37th Int. Conf. on Machine Learning, {ICML}, Proc. of Machine
  Learning Research, vol.~119. {PMLR} (2020)

\bibitem{DBLP:conf/kbse/Abdelkader20}
Abdelkader, H.: Towards robust production machine learning systems: Managing
  dataset shift. In: 35th {IEEE/ACM} Int. Conf. on Automated Software Eng.,
  {ASE} \cite{DBLP:conf/kbse/2020}, pp. 1164--1166

\bibitem{8260754}
{Alagoz}, I., {Hoiss}, T., {German}, R.: Modeling a classifier for solving
  safety-critical binary classification tasks. In: IEEE Int. Conf. on Machine
  Learning and Applications (ICMLA). pp. 914--919 (2017)

\bibitem{DBLP:journals/jss/AnLZCCS20}
An, D., Liu, J., Zhang, M., Chen, X., Chen, M., Sun, H.: Uncertainty modeling
  and runtime verification for autonomous vehicles driving control: {A} machine
  learning-based approach. J. Syst. Softw.  \textbf{167},  110617 (2020)

\bibitem{DBLP:conf/mod/AndersenGG20}
Andersen, P., Goodwin, M., Granmo, O.: Safer reinforcement learning for agents
  in industrial grid-warehousing. In: Nicosia, G., Ojha, V.K., Malfa, E.L.,
  Jansen, G., Sciacca, V., Pardalos, P.M., Giuffrida, G., Umeton, R. (eds.)
  Machine Learning, Optimization, and Data Science - 6th Int. Conf., Revised
  Selected Papers, Part {II}. LNCS, vol. 12566, pp. 169--180. Springer (2020)

\bibitem{ARTEAGA2020104988}
Arteaga, C., Paz, A., Park, J.: Injury severity on traffic crashes: A text
  mining with an interpretable machine-learning approach. Saf. Sci
  \textbf{132},  104988 (2020)

\bibitem{DBLP:conf/icml/AtamturkG20}
Atamt{\"{u}}rk, A., Gomez, A.: Safe screening rules for l0-regression from
  perspective relaxations. In: Proc. of the 37th Int. Conf. on Machine
  Learning, {ICML} \cite{DBLP:conf/icml/2020}, pp. 421--430

\bibitem{DBLP:conf/icml/BaoGH20}
Bao, R., Gu, B., Huang, H.: Fast {OSCAR} and {OWL} regression via safe
  screening rules. In: Proc. of the 37th Int. Conf. on Machine Learning, {ICML}
  \cite{DBLP:conf/icml/2020}, pp. 653--663

\bibitem{DBLP:journals/jss/BarbezKG20}
Barbez, A., Khomh, F., Gu{\'{e}}h{\'{e}}neuc, Y.: A machine-learning based
  ensemble method for anti-patterns detection. J. Syst. Softw.  \textbf{161}
  (2020)

\bibitem{DBLP:conf/nips/BerkenkampTS017}
Berkenkamp, F., Turchetta, M., Schoellig, A.P., Krause, A.: Safe model-based
  reinforcement learning with stability guarantees. In: Advances in Neural
  Info. Processing Syst. 30: Annu. Conf. on Neural Info. Processing Syst. pp.
  908--918 (2017)

\bibitem{10.1007/978-3-030-55583-2_24}
Beyene, T.A., Sahu, A.: Rule-based safety evidence for neural networks. In:
  Casimiro, A., Ortmeier, F., Schoitsch, E., Bitsch, F., Ferreira, P. (eds.)
  Computer Safety, Reliability, and Security. SAFECOMP Workshops. pp. 328--335.
  Springer (2020)

\bibitem{BI2019435}
Bi, X., Shi, X., Zhang, Z.: Cognitive machine learning model for network
  information safety. Sfty. Sci.  \textbf{118},  435 -- 441 (2019)

\bibitem{DBLP:conf/icml/Bou-AmmarTE15}
Bou{-}Ammar, H., Tutunov, R., Eaton, E.: Safe policy search for lifelong
  reinforcement learning with sublinear regret. In: Proc. of the 32nd {ICML}.
  pp. 2361--2369 (2015)

\bibitem{10.1007/978-3-319-99229-7_35}
Bragg, J., Habli, I.: What is acceptably safe for reinforcement learning? In:
  Gallina, B., Skavhaug, A., Schoitsch, E., Bitsch, F. (eds.) Computer Sfty.,
  Reliab., and Sec. pp. 418--430. Springer (2018)

\bibitem{DBLP:journals/jss/BraiekK20}
Braiek, H.B., Khomh, F.: On testing machine learning programs. J. Syst. Softw.
  \textbf{164},  110542 (2020)

\bibitem{DBLP:conf/icml/BrownCSN20}
Brown, D.S., Coleman, R., Srinivasan, R., Niekum, S.: Safe imitation learning
  via fast bayesian reward inference from preferences. In: Proc. of the 37th
  Int. Conf. on Machine Learning, {ICML} \cite{DBLP:conf/icml/2020}, pp.
  1165--1177

\bibitem{10.1007/978-3-319-66284-8_1}
Burton, S., Gauerhof, L., Heinzemann, C.: Making the case for safety of machine
  learning in highly automated driving. In: Tonetta, S., Schoitsch, E., Bitsch,
  F. (eds.) Computer Sfty., Reliab., and Sec. pp. 5--16. Springer (2017)

\bibitem{10.1007/978-3-030-26250-1_30}
Burton, S., Gauerhof, L., Sethy, B.B., Habli, I., Hawkins, R.: Confidence
  arguments for evidence of performance in machine learning for highly
  automated driving functions. In: Romanovsky, A., Troubitsyna, E., Gashi, I.,
  Schoitsch, E., Bitsch, F. (eds.) Computer Sfty., Reliab., and Sec. pp.
  365--377. Springer (2019)

\bibitem{DBLP:conf/nips/ChandakJTWT20}
Chandak, Y., Jordan, S.M., Theocharous, G., White, M., Thomas, P.S.: Towards
  safe policy improvement for non-stationary mdps. In: Larochelle et~al.
  \cite{DBLP:conf/nips/2020}

\bibitem{7838199}
{Chaulwar}, A., {Botsch}, M., {Utschick}, W.: A hybrid machine learning
  approach for planning safe trajectories in complex traffic-scenarios. In:
  IEEE Int. Conf. on Machine Learning and Applications (ICMLA). pp. 540--546
  (2016)

\bibitem{CHEN2019268}
Chen, W., Liu, T., Tang, Y., Xu, D.: Multi-level adaptive coupled method for
  industrial control networks safety based on machine learning. Sfty. Sci.
  \textbf{120},  268 -- 275 (2019)

\bibitem{DBLP:conf/nips/ChowNDG18}
Chow, Y., Nachum, O., Du{\'{e}}{\~{n}}ez{-}Guzm{\'{a}}n, E.A., Ghavamzadeh, M.:
  A lyapunov-based approach to safe reinforcement learning. In: Advances in
  Neural Info. Processing Syst. 31: Annu. Conf. on Neural Info. Processing
  Syst. pp. 8103--8112 (2018)

\bibitem{DBLP:conf/safecomp/CzarneckiS18}
Czarnecki, K., Salay, R.: Towards a framework to manage perceptual uncertainty
  for safe automated driving. In: Gallina et~al.
  \cite{DBLP:conf/safecomp/2018w}, pp. 439--445

\bibitem{DAI201556}
Dai, Y., Tian, J., Rong, H., Zhao, T.: Hybrid safety analysis method based on
  svm and rst: An application to carrier landing of aircraft. Sfty. Sci.
  \textbf{80},  56 -- 65 (2015)

\bibitem{8906396}
{Dhieb}, N., {Ghazzai}, H., {Besbes}, H., {Massoud}, Y.: Extreme gradient
  boosting machine learning algorithm for safe auto insurance operations. In:
  IEEE {ICVES}. pp.~1--5 (2019)

\bibitem{DHULIPALA2020106659}
Dhulipala, S.L., Flint, M.M.: Series of semi-markov processes to model
  infrastructure resilience under multihazards. Reliab. Eng. Syst. Saf.
  \textbf{193},  106659 (2020)

\bibitem{DINDAR201820}
Dindar, S., Kaewunruen, S., An, M., Sussman, J.M.: Bayesian network-based
  probability analysis of train derailments caused by various extreme weather
  patterns on railway turnouts. Sfty. Sci.  \textbf{110},  20 -- 30 (2018),
  railway Sfty.

\bibitem{DUAN2020106676}
Duan, C., Makis, V., Deng, C.: A two-level bayesian early fault detection for
  mechanical equipment subject to dependent failure modes. Reliab. Eng. Syst.
  Saf.  \textbf{193},  106676 (2020)

\bibitem{DBLP:conf/iclr/EysenbachGIL18}
Eysenbach, B., Gu, S., Ibarz, J., Levine, S.: Leave no trace: Learning to reset
  for safe and autonomous reinforcement learning. In: {ICLR}, Proc. (2018)

\bibitem{FAN2020107070}
Fan, S., Blanco-Davis, E., Yang, Z., Zhang, J., Yan, X.: Incorporation of human
  factors into maritime accident analysis using a data-driven bayesian network.
  Reliab. Eng. Syst. Saf.  \textbf{203},  107070 (2020)

\bibitem{FAN2019607}
Fan, Z., Liu, C., Cai, D., Yue, S.: Research on black spot identification of
  safety in urban traffic accidents based on machine learning method. Sfty.
  Sci.  \textbf{118},  607 -- 616 (2019)

\bibitem{FANG2020104604}
Fang, W., Tan, X., Wilbur, D.: Application of intrusion detection technology in
  network safety based on machine learning. Saf. Sci  \textbf{124},  104604
  (2020)

\bibitem{faria2018machine}
Faria, J.M.: Machine learning safety: An overview. In: Proc. of Safety-Critical
  Sys. Symp. (2018)

\bibitem{9307660}
{Ferreira}, R.S.: Towards safety monitoring of ml-based perception tasks of
  autonomous systems. In: IEEE ISSREW. pp. 135--138 (2020)

\bibitem{10.1007/978-3-319-99229-7_48}
Feth, P., Akram, M.N., Schuster, R., Wasenm{\"u}ller, O.: Dynamic risk
  assessment for vehicles of higher automation levels by deep learning. In:
  Gallina, B., Skavhaug, A., Schoitsch, E., Bitsch, F. (eds.) Computer Sfty.,
  Reliab., and Sec. pp. 535--547. Springer (2018)

\bibitem{DBLP:conf/ijcai/FreedBHSB16}
Freed, M., Burns, B., Heller, A., S{\'{a}}nchez, D., Beaumont{-}Bowman, S.: A
  virtual assistant to help dysphagia patients eat safely at home. In: Proc. of
  Int. Joint Conf. on Info., {IJCAI}. pp. 4244--4245 (2016)

\bibitem{DBLP:conf/safecomp/2018w}
Gallina, B., Skavhaug, A., Schoitsch, E., Bitsch, F. (eds.): Computer Sfty.,
  Reliab., and Sec. - {SAFECOMP} Workshops, ASSURE, DECSoS, SASSUR, STRIVE, and
  WAISE, V{\"{a}}ster{\aa}s, Proc., LNCS, vol. 11094. Springer (2018)

\bibitem{10.1007/978-3-030-54549-9_13}
Gauerhof, L., Hawkins, R., Picardi, C., Paterson, C., Hagiwara, Y., Habli, I.:
  Assuring the safety of machine learning for pedestrian detection at
  crossings. In: Casimiro, A., Ortmeier, F., Bitsch, F., Ferreira, P. (eds.)
  Computer Sfty., Reliab., and Sec. pp. 197--212. Springer (2020)

\bibitem{10.1007/978-3-319-99130-6_4}
Gauerhof, L., Munk, P., Burton, S.: Structuring validation targets of a machine
  learning function applied to automated driving. In: Gallina, B., Skavhaug,
  A., Bitsch, F. (eds.) Computer Sfty., Reliab., and Sec. pp. 45--58. Springer
  (2018)

\bibitem{8893310}
{Gharib}, M., {Bondavalli}, A.: On the evaluation measures for machine learning
  algorithms for safety-critical systems. In: European Dependable Computing
  Conf. pp. 141--144 (2019)

\bibitem{DBLP:conf/ijcai/GodoyKGG16}
Godoy, J., Karamouzas, I., Guy, S.J., Gini, M.L.: Moving in a crowd: Safe and
  efficient navigation among heterogeneous agents. In: Proc. of {IJCAI} 2016,
  2016. pp. 294--300 (2016)

\bibitem{DBLP:conf/icml/GuoZJLZ20}
Guo, L., Zhang, Z., Jiang, Y., Li, Y., Zhou, Z.: Safe deep semi-supervised
  learning for unseen-class unlabeled data. In: Proc. of the 37th Int. Conf. on
  Machine Learning, {ICML} \cite{DBLP:conf/icml/2020}, pp. 3897--3906

\bibitem{DBLP:journals/sosym/0001MFT19}
Hartmann, T., Moawad, A., Fouquet, F., Traon, Y.L.: The next evolution of
  {MDE:} a seamless integration of machine learning into domain modeling.
  Softw. Syst. Model.  \textbf{18}(2),  1285--1304 (2019)

\bibitem{DBLP:conf/ijcai/HeinzmannSOSK19}
Heinzmann, L., Shafaei, S., Osman, M.H., Segler, C., Knoll, A.C.: A framework
  for safety violation identification and assessment in autonomous driving. In:
  Proc. of the Workshop on Info. Sfty. co-located w. Int. Joint Conf. on AI,
  AISafety@IJCAI (2019)

\bibitem{DBLP:conf/icse/2020seams}
Honiden, S., Nitto, E.D., Calinescu, R. (eds.): {SEAMS} '20: {IEEE/ACM} 15th
  Int. Symp. on Software Eng. for Adaptive and Self-Managing Systems, Seoul,
  Republic of Korea, 29 June - 3 July, 2020. {ACM} (2020)

\bibitem{DBLP:conf/nips/HuangWP018}
Huang, J., Wu, F., Precup, D., Cai, Y.: Learning safe policies with expert
  guidance. In: Advances in Neural Info. Processing Syst. 31: Annu. Conf. on
  Neural Info. Processing Syst. pp. 9123--9132 (2018)

\bibitem{HUANG2020107220}
Huang, W., Zhang, Y., Kou, X., Yin, D., Mi, R., Li, L.: Railway dangerous goods
  transportation system risk analysis: An interpretive structural modeling and
  bayesian network combining approach. Reliab. Eng. Syst. Saf.  \textbf{204},
  107220 (2020)

\bibitem{IAMSUMANG2018118}
Iamsumang, C., Mosleh, A., Modarres, M.: Monitoring and learning algorithms for
  dynamic hybrid bayesian network in on-line system health management
  applications. Reliab. Eng. Syst. Saf.  \textbf{178},  118 -- 129 (2018)

\bibitem{10.1007/978-3-319-99229-7_2}
Ishikawa, F., Matsuno, Y.: Continuous argument engineering: Tackling
  uncertainty in machine learning based systems. In: Gallina, B., Skavhaug, A.,
  Schoitsch, E., Bitsch, F. (eds.) Computer Sfty., Reliab., and Sec. pp.
  14--21. Springer (2018)

\bibitem{jenn2020identifying}
Jenn, E., Albore, A., Mamalet, F., Flandin, G., Gabreau, C., Delseny, H.,
  Gauffriau, A., Bonnin, H., Alecu, L., Pirard, J., et~al.: Identifying
  challenges to the certification of machine learning for safety critical
  systems. In: Proc. of {ERTS}, Toulouse, France. pp. 29--31 (2020)

\bibitem{JITWASINKUL2016264}
Jitwasinkul, B., Hadikusumo, B.H., Memon, A.Q.: A bayesian belief network model
  of organizational factors for improving safe work behaviors in thai
  construction industry. Sfty. Sci.  \textbf{82},  264 -- 273 (2016)

\bibitem{KALININA2018164}
Kalinina, A., Spada, M., Burgherr, P.: Application of a bayesian hierarchical
  modeling for risk assessment of accidents at hydropower dams. Sfty. Sci.
  \textbf{110},  164 -- 177 (2018)

\bibitem{KHAN2020104858}
Khan, B., Khan, F., Veitch, B.: A dynamic bayesian network model for ship-ice
  collision risk in the arctic waters. Saf. Sci  \textbf{130},  104858 (2020)

\bibitem{10.1145/3243127.3243129}
Khosrowjerdi, H., Meinke, K.: Learning-based testing for autonomous systems
  using spatial and temporal requirements. In: Proc. of the 1st Int. Workshop
  on Machine Learning and SW Eng. in Symbiosis. p. 6–15. MASES 2018, ACM
  (2018)

\bibitem{KIM2020106959}
Kim, J., Shah, A.U.A., Kang, H.G.: Dynamic risk assessment with bayesian
  network and clustering analysis. Reliab. Eng. Syst. Saf.  \textbf{201},
  106959 (2020)

\bibitem{DBLP:conf/icml/KirschnerMHI019}
Kirschner, J., Mutny, M., Hiller, N., Ischebeck, R., Krause, A.: Adaptive and
  safe bayesian optimization in high dimensions via one-dimensional subspaces.
  In: Proc. of the 36th {ICML}. pp. 3429--3438 (2019)

\bibitem{10.1007/978-3-319-99229-7_36}
Kl{\"a}s, M., Vollmer, A.M.: Uncertainty in machine learning applications: A
  practice-driven classification of uncertainty. In: Gallina, B., Skavhaug, A.,
  Schoitsch, E., Bitsch, F. (eds.) Computer Sfty., Reliab., and Sec. pp.
  431--438. Springer (2018)

\bibitem{KONDO2018225}
Kondo, M.C., Morrison, C., Guerra, E., Kaufman, E.J., Wiebe, D.J.: Where do
  bike lanes work best? a bayesian spatial model of bicycle lanes and bicycle
  crashes. Sfty. Sci.  \textbf{103},  225 -- 233 (2018)

\bibitem{DBLP:conf/ijcai/KrakovnaOML19}
Krakovna, V., Orseau, L., Martic, M., Legg, S.: Penalizing side effects using
  stepwise relative reachability. In: Proc. of the Workshop on Info. Sfty.
  co-located w. Int. Joint Conf. on AI, AISafety@IJCAI (2019)

\bibitem{KUMAR2018686}
Kumar, L., Sripada, S.K., Sureka, A., Rath, S.K.: Effective fault prediction
  model developed using least square support vector machine (lssvm). J. of
  Syst. and SW  \textbf{137},  686 -- 712 (2018)

\bibitem{KURIAN2020104873}
Kurian, D., Sattari, F., Lefsrud, L., Ma, Y.: Using machine learning and
  keyword analysis to analyze incidents and reduce risk in oil sands
  operations. Saf. Sci  \textbf{130},  104873 (2020)

\bibitem{DBLP:conf/kbse/Kwiatkowska20}
Kwiatkowska, M.: Safety and robustness for deep learning with provable
  guarantees. In: 35th {IEEE/ACM} Int. Conf. on Automated Software Eng., {ASE}
  \cite{DBLP:conf/kbse/2020}, pp.~1--3

\bibitem{DBLP:conf/nips/2020}
Larochelle, H., Ranzato, M., Hadsell, R., Balcan, M., Lin, H. (eds.): Advances
  in Neural Information Processing Systems 33: Annu. Conf. on Neural
  Information Processing Systems, NeurIPS (2020)

\bibitem{DBLP:conf/mod/LazzeriniP17}
Lazzerini, B., Pistolesi, F.: Artificial bee colony optimization to reallocate
  personnel to tasks improving workplace safety. In: Machine Learning,
  Optimization, and Big Data - Third Int. Conf., {MOD}, Revised Selected
  Papers. pp. 210--221 (2017)

\bibitem{LI2020104655}
Li, F., Chen, C.H., Zheng, P., Feng, S., Xu, G., Khoo, L.P.: An explorative
  context-aware machine learning approach to reducing human fatigue risk of
  traffic control operators. Saf. Sci  \textbf{125},  104655 (2020)

\bibitem{ceabc3b89afc458ab6b979573771641c}
Li, G., Li, Y., Jha, S., Tsai, T., Sullivan, M., Hari, S., Kalbarczyk, Z.,
  Iyer, R.: Av-fuzzer: Finding safety violations in autonomous driving systems.
  In: Vieira, M., Madeira, H., Antunes, N., Zheng, Z. (eds.) Proceedings - IEEE
  31st Int. Symp. on Software Reliab. Eng., ISSRE. pp. 25--36. Proceedings -
  Int. Symp. on Software Reliab. Eng., ISSRE, IEEE Computer Society (Oct 2020)

\bibitem{LIANG2019861}
Liang, X., He, H., Zhang, Y.: Optimization design of micro-piles in landslide
  safety protection based on machine learning. Sfty. Sci.  \textbf{118},  861
  -- 867 (2019)

\bibitem{LIU2019764}
Liu, H., Tian, G.: Building engineering safety risk assessment and early
  warning mechanism construction based on distributed machine learning
  algorithm. Sfty. Sci.  \textbf{120},  764 -- 771 (2019)

\bibitem{DBLP:conf/nips/LuoSK20}
Luo, W., Sun, W., Kapoor, A.: Multi-robot collision avoidance under uncertainty
  with probabilistic safety barrier certificates. In: Larochelle et~al.
  \cite{DBLP:conf/nips/2020}

\bibitem{DBLP:conf/ijcai/MancusoKLS19}
Mancuso, J., Kisielewski, T., Lindner, D., Singh, A.: Detecting spiky
  corruption in markov decision processes. In: Proc. of the Workshop on Info.
  Sfty. co-located w. Int. Joint Conf. on AI, AISafety@IJCAI (2019)

\bibitem{DBLP:conf/ijcai/MansouriLHP19}
Mansouri, M., Lacerda, B., Hawes, N., Pecora, F.: Multi-robot planning under
  uncertain travel times and safety constraints. In: Proc. of {IJCAI}. pp.
  478--484 (2019)

\bibitem{DBLP:journals/jss/MartinVANB20}
Mart{\'{\i}}n, C.L., Villuendas{-}Rey, Y., Azzeh, M., Nassif, A.B., Banitaan,
  S.: Transformed \emph{k}-nearest neighborhood output distance minimization
  for predicting the defect density of software projects. J. Syst. Softw.
  \textbf{167},  110592 (2020)

\bibitem{DBLP:conf/safecomp/MartinelliMNSV18}
Martinelli, F., Mercaldo, F., Nardone, V., Santone, A., Vaglini, G.: Real-time
  driver behaviour characterization through rule-based machine learning. In:
  Gallina et~al.  \cite{DBLP:conf/safecomp/2018w}, pp. 374--386

\bibitem{DBLP:conf/kbse/Masalimov20}
Masalimov, K.A.: A machine learning based approach to autogenerate diagnostic
  models for {CNC} machines. In: 35th {IEEE/ACM} Int. Conf. on Automated
  Software Eng., {ASE} \cite{DBLP:conf/kbse/2020}, pp. 1358--1360

\bibitem{10.1007/978-3-030-26250-1_33}
Matsuno, Y., Ishikawa, F., Tokumoto, S.: Tackling uncertainty in safety
  assurance for machine learning: Continuous argument engineering with
  attributed tests. In: Romanovsky, A., Troubitsyna, E., Gashi, I., Schoitsch,
  E., Bitsch, F. (eds.) Computer Sfty., Reliab., and Sec. pp. 398--404.
  Springer (2019)

\bibitem{8519504}
{Maurya}, S.K., {Choudhary}, A.: Deep learning based vulnerable road user
  detection and collision avoidance. In: IEEE {ICVES}. pp.~1--6 (2018)

\bibitem{DBLP:conf/ijcai/McAllisterGKWSC17}
McAllister, R., Gal, Y., Kendall, A., van~der Wilk, M., Shah, A., Cipolla, R.,
  Weller, A.: Concrete problems for autonomous vehicle safety: Advantages of
  bayesian deep learning. In: Proc. of {IJCAI}. pp. 4745--4753 (2017)

\bibitem{DBLP:conf/ijcai/McDermidJH19}
McDermid, J., Jia, Y., Habli, I.: Towards a framework for safety assurance of
  autonomous systems. In: Proc. of the Workshop on Info. Sfty. co-located w.
  Int. Joint Conf. on AI, AISafety@IJCAI (2019)

\bibitem{MEANGO2020106827}
Meango, T.J.M., Ouali, M.S.: Failure interaction model based on extreme shock
  and markov processes. Reliab. Eng. Syst. Saf.  \textbf{197},  106827 (2020)

\bibitem{DBLP:conf/nips/MhamdiGHM17}
Mhamdi, E.M.E., Guerraoui, R., Hendrikx, H., Maurer, A.: Dynamic safe
  interruptibility for decentralized multi-agent reinforcement learning. In:
  Advances in Neural Info. Processing Syst. 30: Annu. Conf. on Neural Info.
  Processing Syst. pp. 130--140 (2017)

\bibitem{DBLP:journals/jss/MostaeenRRSS20}
Mostaeen, G., Roy, B., Roy, C.K., Schneider, K.A., Svajlenko, J.: A machine
  learning based framework for code clone validation. J. Syst. Softw.
  \textbf{169},  110686 (2020)

\bibitem{OEHLING201989}
Oehling, J., Barry, D.J.: Using machine learning methods in airline flight data
  monitoring to generate new operational safety knowledge from existing data.
  Sfty. Sci.  \textbf{114},  89 -- 104 (2019)

\bibitem{PALTRINIERI2019475}
Paltrinieri, N., Comfort, L., Reniers, G.: Learning about risk: Machine
  learning for risk assessment. Sfty. Sci.  \textbf{118},  475 -- 486 (2019)

\bibitem{DBLP:conf/icse/PasareanuCFG20}
Pasareanu, C.S., Converse, H., Filieri, A., Gopinath, D.: On the probabilistic
  analysis of neural networks. In: Honiden et~al.
  \cite{DBLP:conf/icse/2020seams}, pp.~5--8

\bibitem{PATRIARCA201749}
Patriarca, R., {Di Gravio}, G., Costantino, F.: A monte carlo evolution of the
  functional resonance analysis method (fram) to assess performance variability
  in complex systems. Sfty. Sci.  \textbf{91},  49 -- 60 (2017)

\bibitem{PENG2019263}
Peng, T., Li, C., Zhou, X.: Application of machine learning to laboratory
  safety management assessment. Sfty. Sci.  \textbf{120},  263 -- 267 (2019)

\bibitem{10.1007/978-3-030-26601-1_12}
Picardi, C., Hawkins, R., Paterson, C., Habli, I.: A pattern for arguing the
  assurance of machine learning in medical diagnosis systems. In: Romanovsky,
  A., Troubitsyna, E., Bitsch, F. (eds.) Computer Sfty., Reliab., and Sec. pp.
  165--179. Springer (2019)

\bibitem{DBLP:journals/ress/PiriouFL17}
Piriou, P., Faure, J., Lesage, J.: Generalized boolean logic driven markov
  processes: {A} powerful modeling framework for model-based safety analysis of
  dynamic repairable and reconfigurable systems. Reliab. Eng. Syst. Sfty.
  \textbf{163},  57--68 (2017)

\bibitem{7991914}
{Pop}, D.O., {Rogozan}, A., {Nashashibi}, F., {Bensrhair}, A.: Pedestrian
  recognition through different cross-modality deep learning methods. In: IEEE
  {ICVES}. pp. 133--138 (2017)

\bibitem{DBLP:conf/iclr/RaghunathanSL18}
Raghunathan, A., Steinhardt, J., Liang, P.: Certified defenses against
  adversarial examples. In: {ICLR}, Proc. (2018)

\bibitem{RAHMAN2020106623}
Rahman, M.S., Khan, F., Shaikh, A., Ahmed, S., Imtiaz, S.: A conditional
  dependence-based marine logistics support risk model. Reliab. Eng. Syst. Saf.
   \textbf{193},  106623 (2020)

\bibitem{10.1145/3194133.3194147}
Rodrigues, A., Caldas, R.D., Rodrigues, G.N., Vogel, T., Pelliccione, P.: A
  learning approach to enhance assurances for real-time self-adaptive systems.
  In: Proc. of Int. Conf. on SW Eng. for Adaptive and Self-Managing Systems. p.
  206–216. SEAMS, ACM (2018)

\bibitem{RUIZCASTRO2020106797}
Ruiz-Castro, J.E.: A complex multi-state k-out-of-n: G system with preventive
  maintenance and loss of units. Reliab. Eng. Syst. Saf.  \textbf{197},  106797
  (2020)

\bibitem{8987559}
{Salay}, R., {Angus}, M., {Czarnecki}, K.: A safety analysis method for
  perceptual components in automated driving. In: IEEE Int. Symp. on SW Reliab.
  Eng. (ISSRE). pp. 24--34 (2019)

\bibitem{SARKAR2020104900}
Sarkar, S., Maiti, J.: Machine learning in occupational accident analysis: A
  review using science mapping approach with citation network analysis. Saf.
  Sci  \textbf{131},  104900 (2020)

\bibitem{SARKAR2020104616}
Sarkar, S., Pramanik, A., Maiti, J., Reniers, G.: Predicting and analyzing
  injury severity: A machine learning-based approach using class-imbalanced
  proactive and reactive data. Saf. Sci  \textbf{125},  104616 (2020)

\bibitem{DBLP:conf/icse/ScheererKRB20}
Scheerer, M., Klamroth, J., Reussner, R.H., Beckert, B.: Towards classes of
  architectural dependability assurance for machine-learning-based systems. In:
  Honiden et~al.  \cite{DBLP:conf/icse/2020seams}, pp. 31--37

\bibitem{DBLP:conf/models/SchoneMRA19}
Sch{\"{o}}ne, R., Mey, J., Ren, B., A{\ss}mann, U.: Bridging the gap between
  smart home platforms and machine learning using relational reference
  attribute grammars. In: Burgue{\~{n}}o, L., Pretschner, A., Voss, S.,
  Chaudron, M., Kienzle, J., V{\"{o}}lter, M., G{\'{e}}rard, S., Zahedi, M.,
  Bousse, E., Rensink, A., Polack, F., Engels, G., Kappel, G. (eds.) 22nd
  {ACM/IEEE} Int. Conf. on Model Driven Eng. Languages and Systems Companion,
  {MODELS} Companion. pp. 533--542. {IEEE} (2019)

\bibitem{10.1007/978-3-030-55583-2_29}
Schwalbe, G., Knie, B., S{\"a}mann, T., Dobberphul, T., Gauerhof, L.,
  Raafatnia, S., Rocco, V.: Structuring the safety argumentation for deep
  neural network based perception in automotive applications. In: Casimiro, A.,
  Ortmeier, F., Schoitsch, E., Bitsch, F., Ferreira, P. (eds.) Computer Safety,
  Reliability, and Security. SAFECOMP Workshops. pp. 383--394. Springer (2020)

\bibitem{10.1007/978-3-319-99229-7_39}
Shafaei, S., Kugele, S., Osman, M.H., Knoll, A.: Uncertainty in machine
  learning: A safety perspective on autonomous driving. In: Gallina, B.,
  Skavhaug, A., Schoitsch, E., Bitsch, F. (eds.) Computer Sfty., Reliab., and
  Sec. pp. 458--464. Springer (2018)

\bibitem{7990255}
{Shepperd}, M., {Hall}, T., {Bowes}, D.: Authors’ reply to “comments on
  ‘researcher bias: The use of machine learning in software defect
  prediction’”. IEEE Trans. on SW Eng.  \textbf{44}(11),  1129--1131 (2018)

\bibitem{DBLP:conf/iclr/SinghGPV19}
Singh, G., Gehr, T., P{\"{u}}schel, M., Vechev, M.T.: Boosting robustness
  certification of neural networks. In: {ICLR} (2019)

\bibitem{DBLP:conf/iclr/SinhaND18}
Sinha, A., Namkoong, H., Duchi, J.C.: Certifying some distributional robustness
  with principled adversarial training. In: {ICLR}, Proc. (2018)

\bibitem{DBLP:conf/nips/SinhaOTD20}
Sinha, A., O'Kelly, M., Tedrake, R., Duchi, J.C.: Neural bridge sampling for
  evaluating safety-critical autonomous systems. In: Larochelle et~al.
  \cite{DBLP:conf/nips/2020}

\bibitem{8359087}
{Song}, Q., {Guo}, Y., {Shepperd}, M.: A comprehensive investigation of the
  role of imbalanced learning for software defect prediction. IEEE Trans. on SW
  Eng.  \textbf{45}(12),  1253--1269 (2019)

\bibitem{STEIJN2020104514}
Steijn, W., {Van Kampen}, J., {Van der Beek}, D., Groeneweg, J., {Van Gelder},
  P.: An integration of human factors into quantitative risk analysis using
  bayesian belief networks towards developing a ‘qra+’. Saf. Sci
  \textbf{122},  104514 (2020)

\bibitem{DBLP:conf/ijcai/SternJ17}
Stern, R., Juba, B.: Efficient, safe, and probably approximately complete
  learning of action models. In: Proc. of {IJCAI}. pp. 4405--4411 (2017)

\bibitem{DBLP:conf/icml/SuiGBK15}
Sui, Y., Gotovos, A., Burdick, J.W., Krause, A.: Safe exploration for
  optimization with gaussian processes. In: Proc. of the 32nd {ICML}. pp.
  997--1005 (2015)

\bibitem{DBLP:conf/icml/SunDK17}
Sun, W., Dey, D., Kapoor, A.: Safety-aware algorithms for adversarial
  contextual bandit. In: Proc. of {ICML}. pp. 3280--3288 (2017)

\bibitem{TALEBBERROUANE2020106587}
Taleb-Berrouane, M., Khan, F., Amyotte, P.: Bayesian stochastic petri nets
  (bspn) - a new modelling tool for dynamic safety and reliability analysis.
  Reliab. Eng. Syst. Saf.  \textbf{193},  106587 (2020)

\bibitem{7450669}
{Tantithamthavorn}, C., {McIntosh}, S., {Hassan}, A.E., {Matsumoto}, K.:
  Comments on “researcher bias: The use of machine learning in software
  defect prediction”. IEEE Trans. on SW Eng.  \textbf{42}(11),  1092--1094
  (2016)

\bibitem{DBLP:journals/tse/Tantithamthavorn16}
Tantithamthavorn, C., McIntosh, S., Hassan, A.E., Matsumoto, K.: Comments on
  "researcher bias: The use of machine learning in software defect prediction".
  {IEEE} Trans. SW Eng.  \textbf{42}(11),  1092--1094 (2016)

\bibitem{DBLP:conf/nips/TsuzukuSS18}
Tsuzuku, Y., Sato, I., Sugiyama, M.: Lipschitz-margin training: Scalable
  certification of perturbation invariance for deep neural networks. In:
  Advances in Neural Info. Processing Syst. 31: Annu. Conf. on Neural Info.
  Processing Syst., NeurIPS. pp. 6542--6551 (2018)

\bibitem{DBLP:conf/nips/TurchettaB016}
Turchetta, M., Berkenkamp, F., Krause, A.: Safe exploration in finite markov
  decision processes with gaussian processes. In: Advances in Neural Info.
  Processing Syst. 29: Annu. Conf. on Neural Info. Processing Syst. pp.
  4305--4313 (2016)

\bibitem{DBLP:conf/nips/TurchettaKS0A20}
Turchetta, M., Kolobov, A., Shah, S., Krause, A., Agarwal, A.: Safe
  reinforcement learning via curriculum induction. In: Larochelle et~al.
  \cite{DBLP:conf/nips/2020}

\bibitem{DBLP:conf/ijcai/TurnerHT19}
Turner, A.M., Hadfield{-}Menell, D., Tadepalli, P.: Conservative agency. In:
  Proc. of the Workshop on Info. Sfty. co-located w. Int. Joint Conf. on AI,
  AISafety@IJCAI (2019)

\bibitem{UTNE2020106757}
Utne, I.B., Rokseth, B., Sørensen, A.J., Vinnem, J.E.: Towards supervisory
  risk control of autonomous ships. Reliab. Eng. Syst. Saf.  \textbf{196},
  106757 (2020)

\bibitem{DBLP:conf/nips/WangPWYJ18}
Wang, S., Pei, K., Whitehouse, J., Yang, J., Jana, S.: Efficient formal safety
  analysis of neural networks. In: Advances in Neural Info. Processing Syst.
  31: Annu. Conf. on Neural Info. Processing Syst. pp. 6369--6379 (2018)

\bibitem{DBLP:journals/ase/WangZJZ16}
Wang, T., Zhang, Z., Jing, X., Zhang, L.: Multiple kernel ensemble learning for
  software defect prediction. Autom. Softw. Eng.  \textbf{23}(4),  569--590
  (2016)

\bibitem{WANG2019833}
qiu Wang, X., Yin, J.: Application of machine learning in safety evaluation of
  athletes training based on physiological index monitoring. Sfty. Sci.
  \textbf{120},  833 -- 837 (2019)

\bibitem{WANG2020106705}
Wang, X., Zhao, X., Wang, S., Sun, L.: Reliability and maintenance for
  performance-balanced systems operating in a shock environment. Reliab. Eng.
  Syst. Saf.  \textbf{195},  106705 (2020)

\bibitem{WANG2020106781}
Wang, Z., Li, S.: Data-driven risk assessment on urban pipeline network based
  on a cluster model. Reliab. Eng. Syst. Saf.  \textbf{196},  106781 (2020)

\bibitem{10.1007/978-3-030-55583-2_30}
Ward, F.R., Habli, I.: An assurance case pattern for the interpretability of
  machine learning in safety-critical systems. In: Casimiro, A., Ortmeier, F.,
  Schoitsch, E., Bitsch, F., Ferreira, P. (eds.) Computer Safety, Reliability,
  and Security. SAFECOMP Workshops. pp. 395--407. Springer (2020)

\bibitem{WASHINGTON2019654}
Washington, A., Clothier, R., Neogi, N., Silva, J., Hayhurst, K., Williams, B.:
  Adoption of a bayesian belief network for the system safety assessment of
  remotely piloted aircraft systems. Sfty. Sci.  \textbf{118},  654 -- 673
  (2019)

\bibitem{DBLP:conf/nips/WenT18}
Wen, M., Topcu, U.: Constrained cross-entropy method for safe reinforcement
  learning. In: Advances in Neural Info. Processing Syst. 31: Annu. Conf. on
  Neural Info. Processing Syst. pp. 7461--7471 (2018)

\bibitem{10.1007/978-3-030-55583-2_25}
Willers, O., Sudholt, S., Raafatnia, S., Abrecht, S.: Safety concerns and
  mitigation approaches regarding the use of deep learning in safety-critical
  perception tasks. In: Casimiro, A., Ortmeier, F., Schoitsch, E., Bitsch, F.,
  Ferreira, P. (eds.) Computer Safety, Reliability, and Security. SAFECOMP
  Workshops. pp. 336--350. Springer (2020)

\bibitem{DBLP:conf/ijcai/Wotawa19}
Wotawa, F.: On the importance of system testing for assuring safety of {AI}
  systems. In: Proc. of the Workshop on Info. Sfty. co-located w. Int. Joint
  Conf. on AI, AISafety@IJCAI (2019)

\bibitem{10.1007/978-3-030-55583-2_28}
Wozniak, E., C{\^a}rlan, C., Acar-Celik, E., Putzer, H.J.: A safety case
  pattern for systems with machine learning components. In: Casimiro, A.,
  Ortmeier, F., Schoitsch, E., Bitsch, F., Ferreira, P. (eds.) Computer Safety,
  Reliability, and Security. SAFECOMP Workshops. pp. 370--382. Springer (2020)

\bibitem{WU2020107051}
Wu, B., Cui, L.: Reliability evaluation of markov renewal shock models with
  multiple failure mechanisms. Reliab. Eng. Syst. Saf.  \textbf{202},  107051
  (2020)

\bibitem{WU2015157}
Wu, X., Liu, H., Zhang, L., Skibniewski, M.J., Deng, Q., Teng, J.: A dynamic
  bayesian network based approach to safety decision support in tunnel
  construction. Reliab. Eng. Syst. Saf.  \textbf{134},  157 -- 168 (2015)

\bibitem{DBLP:journals/corr/abs-2008-08221}
Xu, Z., Saleh, J.H.: Machine learning for reliability eng. and safety
  applications: Review of current status and future opportunities. CoRR
  \textbf{abs/2008.08221} (2020)

\bibitem{XU2020107210}
Xu, Z., Saleh, J.H., Subagia, R.: Machine learning for helicopter accident
  analysis using supervised classification: Inference, prediction, and
  implications. Reliab. Eng. Syst. Saf.  \textbf{204},  107210 (2020)

\bibitem{YANG2019205}
Yang, B.: Dynamic risk identification safety model based on fuzzy support
  vector machine and immune optimization algorithm. Sfty. Sci.  \textbf{118},
  205 -- 211 (2019)

\bibitem{YANG2020104437}
Yang, B.: Construction of logistics financial security risk ontology model
  based on risk association and machine learning. Saf. Sci  \textbf{123},
  104437 (2020)

\bibitem{YU2019248}
Yu, Q., Zhou, Y.: Traffic safety analysis on mixed traffic flows at signalized
  intersection based on haar-adaboost algorithm and machine learning. Sfty.
  Sci.  \textbf{120},  248 -- 253 (2019)

\bibitem{YU201744}
Yu, X., Liu, J., Yang, Z., Liu, X.: The bayesian network based program
  dependence graph and its application to fault localization. J. of Syst. and
  SW  \textbf{134},  44 -- 53 (2017)

\bibitem{ZHANG2020107094}
Zhang, N., Si, W.: Deep reinforcement learning for condition-based maintenance
  planning of multi-component systems under dependent competing risks. Reliab.
  Eng. Syst. Saf.  \textbf{203},  107094 (2020)

\bibitem{DBLP:conf/ijcai/ZhangYCAB16}
Zhang, R., Yu, Y., Chamie, M.E., A{\c{c}}ikmese, B., Ballard, D.H.:
  Decision-making policies for heterogeneous autonomous multi-agent systems
  with safety constraints. In: Proc. {IJCAI}. pp. 546--553 (2016)

\bibitem{DBLP:conf/ijcai/ZhangDS18}
Zhang, S., Durfee, E.H., Singh, S.P.: Minimax-regret querying on side effects
  for safe optimality in factored markov decision processes. In: Proc. of
  {IJCAI}. pp. 4867--4873 (2018)

\bibitem{ZHANG2020107089}
Zhang, Y., Weng, W.: Bayesian network model for buried gas pipeline failure
  analysis caused by corrosion and external interference. Reliab. Eng. Syst.
  Saf.  \textbf{203},  107089 (2020)

\bibitem{DBLP:journals/ase/ZhangJW17}
Zhang, Z., Jing, X., Wang, T.: Label propagation based semi-supervised learning
  for software defect prediction. Autom. Softw. Eng.  \textbf{24}(1),  47--69
  (2017)

\bibitem{ZHAO2020104888}
Zhao, D., Wang, Z., Song, Z., Guo, P., Cao, X.: Assessment of domino effects in
  the coal gasification process using {Fuzzy Analytic Hierarchy Process and
  Bayesian Network}. Saf. Sci  \textbf{130},  104888 (2020)

\bibitem{DBLP:conf/safecomp/ZhaoBSRF0020}
Zhao, X., Banks, A., Sharp, J., Robu, V., Flynn, D., Fisher, M., Huang, X.: A
  safety framework for critical systems utilising deep neural networks. In:
  Casimiro, A., Ortmeier, F., Bitsch, F., Ferreira, P. (eds.) Computer Safety,
  Reliability, and Security - 39th Int. Conf., {SAFECOMP} Proc. LNCS, vol.
  12234, pp. 244--259. Springer (2020)

\bibitem{ZHAO2020106878}
Zhao, Y., Huang, L., Smidts, C., Zhu, Q.: Finite-horizon semi-markov game for
  time-sensitive attack response and probabilistic risk assessment in nuclear
  power plants. Reliab. Eng. Syst. Saf.  \textbf{201},  106878 (2020)

\bibitem{ZHAO201774}
Zhao, Z., {Bin Liang}, Wang, X., Lu, W.: Remaining useful life prediction of
  aircraft engine based on degradation pattern learning. Reliab. Eng. Syst.
  Saf.  \textbf{164},  74 -- 83 (2017)

\bibitem{ZHOU2020106902}
Zhou, T., Peng, Y.: Adaptive bayesian quadrature based statistical moments
  estimation for structural reliability analysis. Reliab. Eng. Syst. Saf.
  \textbf{198},  106902 (2020)

\bibitem{ZHOU2018152}
Zhou, Y., Li, C., Zhou, C., Luo, H.: Using bayesian network for safety risk
  analysis of diaphragm wall deflection based on field data. Reliab. Eng. Syst.
  Saf.  \textbf{180},  152 -- 167 (2018)

\bibitem{DBLP:conf/nips/ZimmerMN18}
Zimmer, C., Meister, M., Nguyen{-}Tuong, D.: Safe active learning for
  time-series modeling with gaussian processes. In: Advances in Neural Info.
  Processing Syst. 31: Annu. Conf. on Neural Info. Processing Syst. pp.
  2735--2744 (2018)

\end{thebibliography}
\end{document}